\documentclass[letterpaper, 10 pt, conference]{ieeeconf}  

\IEEEoverridecommandlockouts                              

\usepackage{graphics} 
\usepackage{epsfig} 
\usepackage{mathptmx} 
\usepackage{times} 
\usepackage{amsmath} 
\usepackage{amssymb}  

\usepackage{flushend}

\title{\LARGE \bf
One Body, Two Minds: Variable Autonomy Approach for a Co-embodied Robotic Hand 
}

\author{
	Piotr Koczy$^{1}$,
	Yuchong Zhang$^{1,\ast}$,
	Danica Kragic$^{1}$,
    Michael C. Welle$^{1,2,\ast}$
\thanks{$^{1}$KTH Royal Institute of Technology, Stockholm, Sweden.
        {\tt\small pkoczy, yuchongz, dani, mwelle@kth.se}}%
\thanks{$^{2}$INCAR Robotics AB, Stockholm, Sweden.
        {\tt\small michael.welle@incar-robotics.se}}%
\thanks{$^{\ast}$Corresponding authors.}
}

\begin{document}

\maketitle
\thispagestyle{empty}
\pagestyle{empty}

\begin{abstract}

Assistive robotic systems face a fundamental trade-off: fully autonomous systems lack user agency, while fully user-controlled systems demand continuous cognitive effort. Existing shared autonomy approaches blend human and robot commands but are mostly deployed in separate physical bodies. We introduce co-embodiment with variable autonomy, where human and robot share a single physical body and operate at different autonomy levels across task phases, from mutual autonomy during object search and grasping to human-dominant control during actuation.

We present a co-embodied, wearable robotic hand that has it's own "mind" and operates with variable autonomy levels. A learning-from-demonstration visuomotor diffusion policy enables autonomous grasping when the user positions the hand near known objects. Once grasped, the system signals completion and the human can actuate the grasped tool (drill, spray bottle, infrared thermometer, lighter, ice-cream scoop) via hands-free head gestures. The human retains veto authority 
at all times through a release gesture that returns the system to the initial phase. Unlike blended autonomy, where control is continuously negotiated, our co-embodied approach consists of variable autonomy from full human control to full independent actions while maintaining physical coupling, realizing a one body two minds paradigm.

In a user study with 44 participants performing five bimanual tasks, users rapidly adapted to this "two minds" paradigm: completion times improved $23.3\%$ across trials ($p<0.001$, Cohen's $d=0.94$), the best-performing policy variant reached $93.6\%$ task success rate, and acceptance ratings were high ($5.70/7$ overall impression, $5.52/7$ daily use willingness). This work establishes co-embodiment with variable autonomy as a viable approach 
for assistive robotics, enabling human-robot collaboration through co-embodiment.

\end{abstract}

\section{INTRODUCTION}
\noindent
The human hand is a uniquely dexterous instrument, enabling complex interactions with the world, from delicate grasps to high-force tool use. Robotic systems that aim to replicate or enhance these capabilities often face a difficult trade-off: they are either fully autonomous or entirely user-controlled. In the realm of prosthetics and wearable robotics, this dichotomy limits the potential for intuitive, seamless assistance during real-world tasks. What if, instead, robotic and human control could collaborate fluidly, each having its own "mind" while being physically coupled, varying autonomy levell when best suited.

Upper-limb impairment affects millions of people worldwide, with stroke alone accounting for substantial loss of arm/hand functionality annually \cite{gbd2021global}. Bimanual tasks, such as cooking, dressing, assembling objects, personal care, and using tools, form the foundation of independent living, yet remain profoundly challenging with single-handed function~\cite{sleimen2011bimanual}. While robotic technologies promise to restore this capability, current approaches remain constrained by their underlying control and integration paradigms.

\noindent
\textbf{The Spectrum of Human-Robot Control:} Existing assistive systems occupy distinct points along a control spectrum.
At one extreme, fully autonomous robotic assistants operate independently, fetching objects or manipulating tools without much direct user input~\cite{park2020active}. These systems excel at structured tasks but lack adaptability to user intent and remove human agency, which is critical for acceptance and practical deployment~\cite{methnani2024s}. At the other extreme, manually controlled prosthetics require users to consciously direct every movement, typically through myoelectric signals~\cite{geethanjali2016myoelectric}, mechanical interfaces~\cite{brown2016empirical}, or neural control~\cite{ortiz2023highly}. While preserving user agency, this approach demands sustained cognitive effort and struggles with the complexity of dexterous manipulation~\cite{schultz2011neural}.
Between these extremes lies shared autonomy, where human and robot continuously negotiate control through blended commands~\cite{dragan2013policy}. Successful implementations in wheelchair navigation~\cite{chang2017shared}, robotic arms~\cite{jonnavittula2024sari}, and teleoperation~\cite{javdani2018shared} demonstrate the power of this approach. 
However, shared autonomy systems typically maintain physical separation between human and robot, as the user operates a joystick to guide a separate robotic entity or adjusts assistance levels for a wheelchair. This separation simplifies the control problem but limits the directness of physical interaction and situational awareness that comes from bodily integration.

\noindent
\textbf{The Missing Paradigm of Co-Embodiment:} A fundamentally different approach remains largely unexplored: \textit{co-embodiment with variable autonomy}, where human and robot share a single physical body and operate at different autonomy levels based on task phase. This paradigm differs from shared autonomy in two critical ways. 
\emph{i)} first, the physical coupling means the human directly positions and moves the robotic system through their own body, simplifying the proprioceptive and spatial awareness challenges. 
\emph{ii)} second, rather than continuously blending commands, autonomy levels vary across task phases, from mutual autonomy where both minds operate independently but coupled through the shared body, to human-dominant phases where the robot provides support while the human retains full control authority.

This variable autonomy structure offers potential advantages for both system design and user experience (UX). From a technical perspective, it simplifies the arbitration problem; instead of continuously resolving conflicting human and robot intentions, the system operates with phase-appropriate autonomy levels.
From a human-factors perspective, this can support clearer mental models: users know when the system must cooperate with an autonomous agenda and when the robot plays a purely supportive/complementary role. Moreover, because the human is the more capable partner, they can compensate for model quirks that might cause failures in a fully autonomous setting, without becoming cognitively overloaded by having to issue low-level commands.

\noindent
\textbf{Core Insight:} Consider a prototypical bimanual task: using a power drill while securing the workpiece. 
The task naturally decomposes into phases requiring different capabilities. Positioning the hand near the drill handle requires human spatial reasoning, understanding where the drill is in the workspace, planning a collision-free path, and adapting to dynamic environments. Grasping the drill demands precision in order to achieve proper finger placement, appropriate grip force, and stable contact geometry. Actuation requires the human to trigger the drill at the right moment and location. Release returns the hand to a neutral state for the next interaction.
Human spatial awareness and high-level planning excel at positioning and task timing, while robotic systems can achieve consistent, reliable grasps through learned policies~\cite{mahler2018dex}. 
Rather than forcing one agent to handle all phases or continuously blending their contributions, co-embodied variable autonomy assigns different autonomy levels to each phase based on the capabilities required. The human positions the shared hand near target objects and decides when to use tools; the robot autonomously executes grasps when objects enter its reach. This division of labor leverages the complementary strengths of human cognition and robotic precision. To guarantee a secure workflow and increase trust in the system, the human always has the option to open the hand at any state.

\noindent
\textbf{Technical Challenge - Autonomous Grasping in the Wild:} 
Enabling robust autonomous grasping for diverse objects in unstructured environments requires learning-based approaches that can generalize beyond programmed behaviors~\cite{chi2025diffusion}. Recent advances in learning from demonstration (LfD) visuomotor policy learning, particularly diffusion-based methods~\cite{chi2025diffusion, janner2022planning, ze20243d, yang2025s}, have demonstrated impressive dexterity on complex manipulation tasks~\cite{ingelhag2024robotic, zhao2024aloha, chi2024universal,barreiros2025careful}, including In-hand manipulation using multifingered hands~\cite{koczy2025learning}. 
These policies can learn multimodal action distributions from expert demonstrations, capture fine-grained sensorimotor coordination, and generalize to similar scenarios. 
However, deploying such policies in a co-embodied system in the wild presents unique challenges: the policy must remain stable when the human moves the shared body, be robust to background changes, and transition smoothly and clearly between active grasping and human-intended triggering of object states.

\noindent
\textbf{Our Contribution:} 
We present, to the best of our knowledge, the first robotic system demonstrating co-embodiment with variable autonomy for assistive bimanual tasks. Our wearable robotic hand integrates with the user's arm, creating a ``one body, two minds'' paradigm where human and robot operate at different autonomy levels across task phases. A visuomotor diffusion policy trained on expert human demonstrations enables autonomous grasping when the user positions the hand near known objects. Once grasped, the system provides audible feedback and enables head gesture commands for tool actuation (nod) and release (shake). Critically, the human's positioning directly influences autonomous grasping performance, creating implicit collaboration leading to a co-embodiment paradigm with one body and two minds.

\begin{figure}[htbp]
    \centering
    \includegraphics[width=\columnwidth]{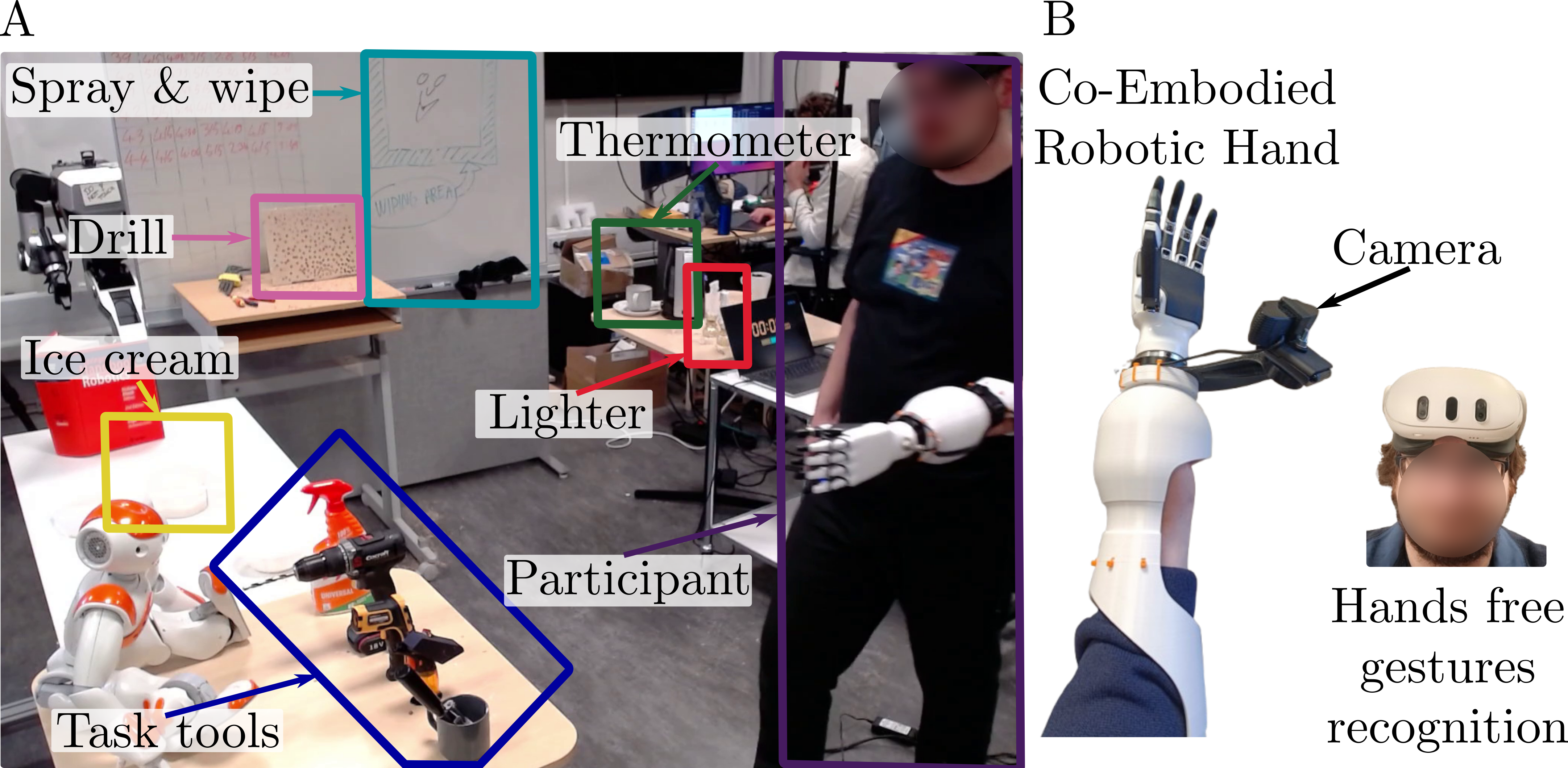}
    \caption{\textbf{Overview of the co-embodied task environment and wearable system.}
    \textbf{(Left)} Experimental workspace with the five task tools used in the evaluation (drill, spray-and-wipe target, thermometer, lighter, and ice cream setup).
    \textbf{(Right)} Co-embodied robotic hand worn on the participant’s forearm with a head-worn headset used for hands-free gestures.
    See Materials and Methods for hardware details, task definitions, and success criteria.}
    \label{fig:setup}
\end{figure}

We evaluate this system with $n=44$ participants performing five bimanual tasks (\emph{ice cream scooping, drilling a hole into a wooden board, spray-and-wipe, measuring tea water with an infrared thermometer, and lighting candles}) across three successive trials following brief familiarization (Fig.~\ref{fig:setup}). We trained three policy variants with different background augmentation strategies (Models A--C) to investigate the relative contributions of user adaptation versus policy robustness to system performance. Primary outcome measures included task completion time, binary task success, and post-study questionnaires assessing usability, acceptance, and embodiment. Extended practice sessions with four highly system-familiar users examined continued skill development potential. Full user study design, experimental procedures, system architecture, and scoring criteria are reported in Materials and Methods and supplamentary material S1, including a summery video M1. Our findings demonstrate that users rapidly adapt to variable autonomy ($23\%$ improvement, $p<0.001$), achieve high success rates ($87\%$ overall), and report strong acceptance (M=$5.48/7$), establishing co-embodiment as a viable approach for assistive robotics.

\section*{Results}
We report objective performance and UX outcomes from the three-trial study. Learnability was quantified within participants using total completion time and per-trial task success (0--5) across repeated trials (Fig.~\ref{fig:learning_effects}, Fig.~\ref{fig:task_success}). To isolate user adaptation from autonomous policy robustness, we compared the same metrics across the three policy variants (Models A--C) under counterbalanced model order (Fig.~\ref{fig:model_comparison}), and we further summarize post-study UX ratings (Fig.~\ref{fig:user_experience}) and extended-practice performance (Fig.~\ref{fig:extended_practice}).

\subsection*{Rapid user adaptation demonstrates learning effect dominance}
\begin{figure}[htbp]
    \centering
    \includegraphics[width=.8\columnwidth]{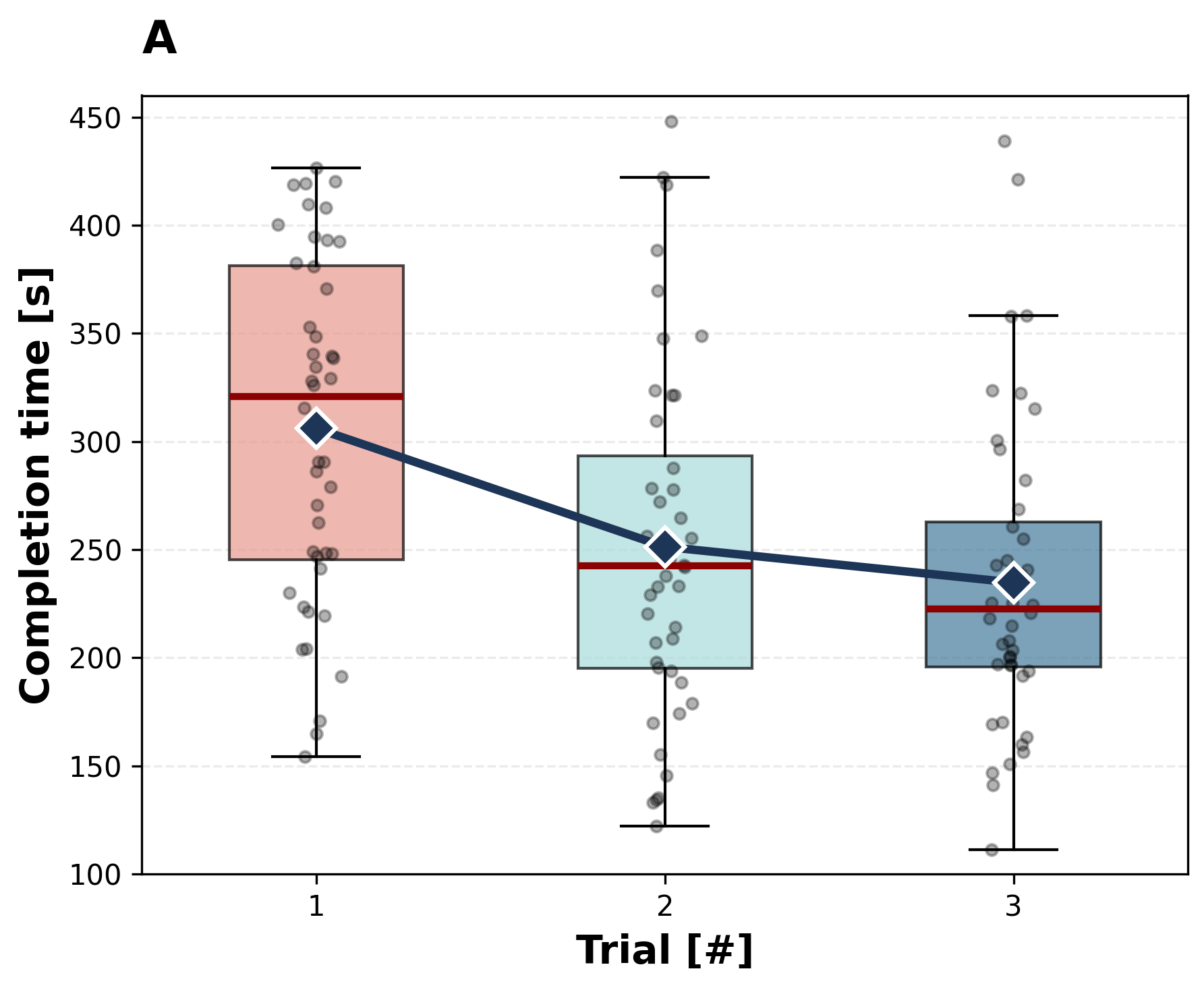}
    \includegraphics[width=.8\columnwidth]{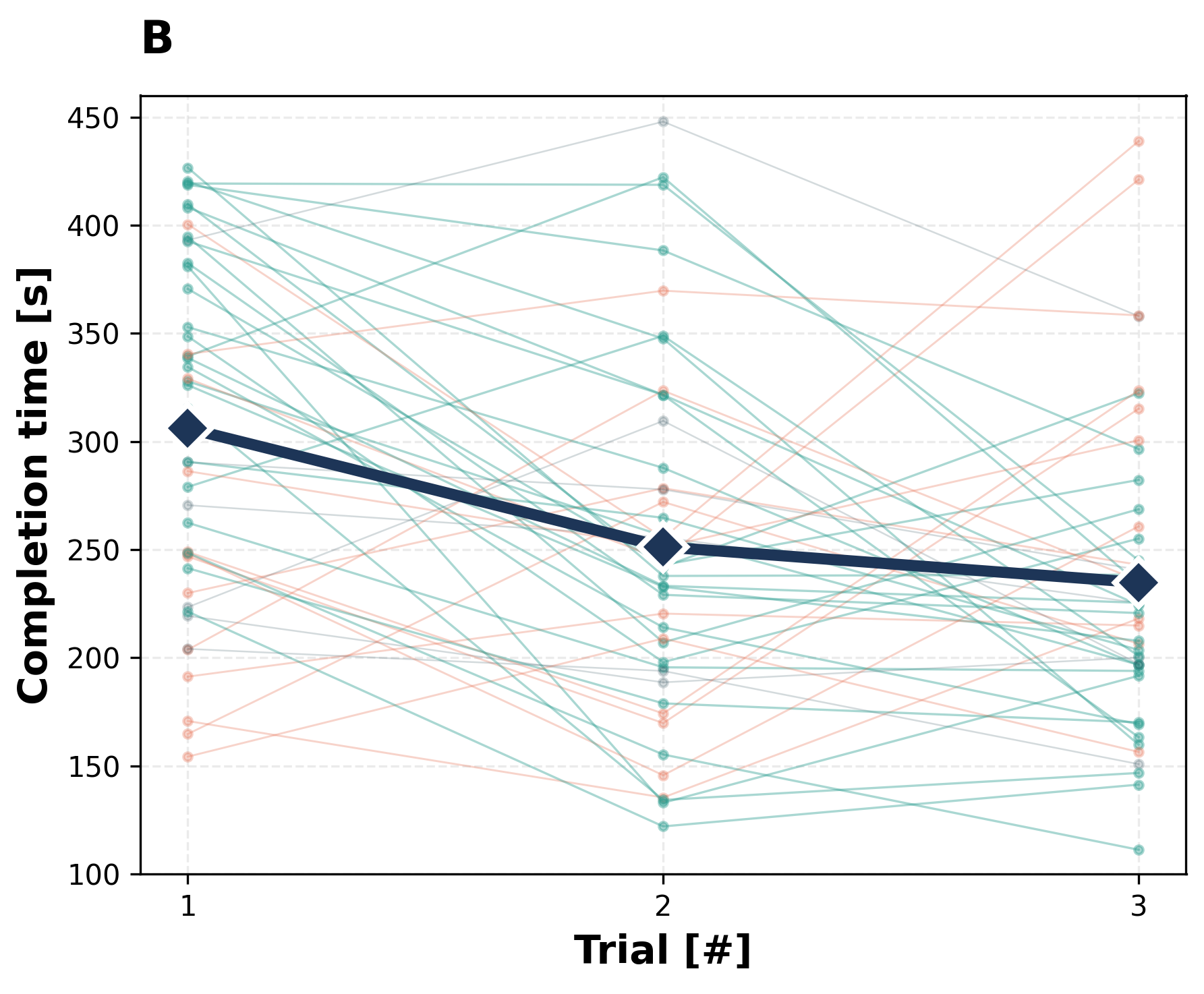}
    \caption{\textbf{Users' completion over the three trials.} We observe rapid user adaptation to co-embodied variable autonomy. (A) Completion times decreased significantly across three trials ($n=44$, Friedman test $p<0.001$). Box plots show median (thick red line), interquartile range (box), and range (whiskers). Individual data points (gray circles) show raw data. Mean trajectory (navy diamonds connected by line) demonstrates $23.3\%$ improvement from Trial $1$ to Trial $3$. (B) Individual learning trajectories reveal heterogeneous patterns. Each thin line represents one participant's performance across trials. Line colors indicate learning magnitude: teal lines show fast learners ($>60$s improvement), gray lines show moderate learners ($0$-$60$s improvement), and orange lines show non-learners (no improvement or decline). The thick navy line shows the group mean trajectory. 
    }
    \label{fig:learning_effects}
\end{figure}

Participants showed significant performance improvement across the three trials, with completion times decreasing from $M=306.1$s ($SD=80.2$s) in Trial $1$ to $M=251.3$s ($SD=81.6$s) in Trial $2$ and $M=234.8$s ($SD=71.5$s) in Trial $3$ (Fig.~\ref{fig:learning_effects}A). This represents a $23.3\%$ improvement from first to final trial. Friedman's test confirmed a significant effect of trial number ($\chi^2=15.95$, $p<0.001$). Post-hoc pairwise comparisons using Wilcoxon signed-rank tests revealed significant improvements between Trials $1$ and $2$ ($W=187.0$, $p<0.001$) and Trials $1$ and $3$ ($W=131.0$, $p<0.0001$), but not between Trials $2$ and $3$ ($W=377.0$, $p=0.172$), indicating users reached plateau performance by the second trial. The effect size for Trial $1$ vs Trial $3$ improvement was large (Cohen's $d=0.94$), demonstrating substantial practical significance.

Individual learning trajectories (Fig.~\ref{fig:learning_effects}B) revealed heterogeneous but predominantly positive patterns, with $70.5\%$ of participants showing overall improvement from Trial $1$ to Trial $3$. The rapid learning curve, with a potential plateau achieved by the second trial, suggests that users quickly developed effective strategies for coordinating with the co-embodied agent while maintaining control over positioning, actuation, and release decisions.

\subsection*{High task success with task-specific challenges}
\begin{figure}[htbp]
    \centering
    \includegraphics[width=.8\columnwidth]{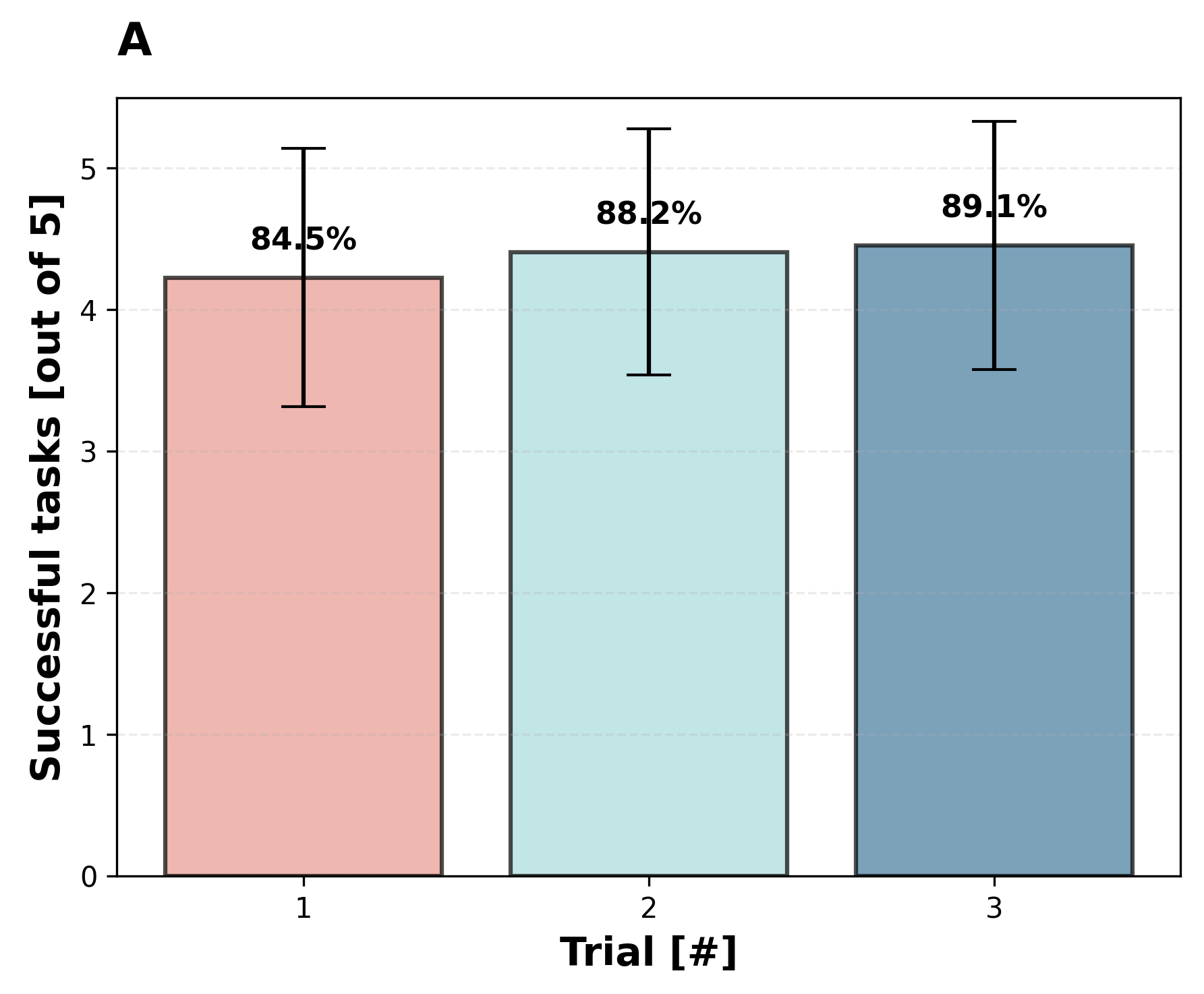}
    \includegraphics[width=.8\columnwidth]{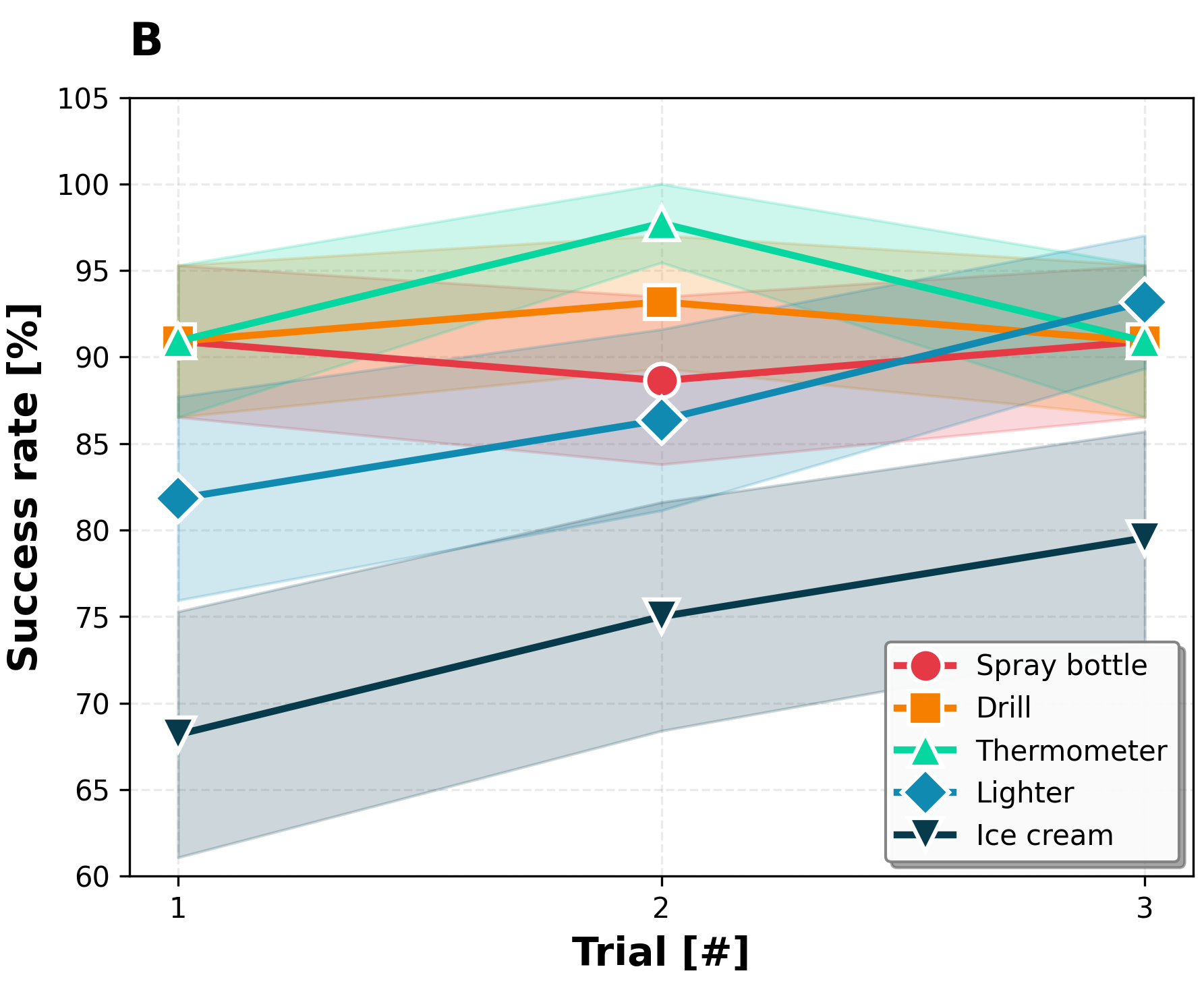}
    \caption{ \textbf{Task success rates across trials.} (A) Overall task success showed modest improvement from Trial $1$ (M = 4.23/5, $84.5\%$) to Trial $3$ (M = 4.45/5, $89.1\%$). Bars show mean successful tasks out of $5$, error bars indicate standard deviation. The improvement was not statistically significant (Friedman test: $\chi^2 = 2.67$, $p = 0.264$). (B) Task-specific learning trajectories reveal differential difficulty. Lines show mean success rates across trials and shaded regions indicating $\pm 1$ standard error. Four tasks maintained high success rates ($\sim 91\%$) throughout all trials, demonstrating ceiling effects. Ice cream (dark blue, downward triangles) showed the steepest learning curve, improving from $68.2\%$ to $79.5\%$ ($+11.4\%$), indicating this was the most challenging task that benefited most from practice. 
    }
    \label{fig:task_success}
\end{figure}

Overall task success rates improved modestly from $84.5\%$ ($M=4.23/5$ tasks, $SD=0.91$) in Trial $1$ to $88.2\%$ ($M=4.41/5$, $SD=0.87$) in Trial $2$ and $89.1\%$ ($M=4.45/5$, $SD=0.87$) in Trial $3$ (Fig.~\ref{fig:task_success}A). This improvement trend, while positive, did not reach statistical significance (Friedman test: $\chi^2=2.67$, $p=0.264$), likely due to ceiling effects with most tasks already highly successful in Trial $1$.
Analysis of individual task success rates revealed distinct difficulty levels for the different tasks (Fig.~\ref{fig:task_success}B). By Trial $3$, the spray bottle ($90.9\%$), drill ($90.9\%$), thermometer ($90.9\%$), and lighter ($93.2\%$) tasks all exceeded $90\%$ success. Two tasks showed notable learning trajectories: the lighter task improved from $81.8\%$ in Trial $1$ to $93.2\%$ in Trial $3$ ($+11.4\%$), and the ice cream scooping task, which proved most challenging, improved from $68.2\%$ (Trial $1$) to $75.0\%$ (Trial $2$) to $79.5\%$ (Trial $3$) ($+11.4\%$) but remained below other tasks. The ice cream task required the most unusual grasping angle as the ice cream scoop was designed for right-handed users.
The combination of significant speed improvement with non-significant success rate change indicates that users became more efficient while maintaining task accuracy.

\subsection*{Model comparison suggests a possible reliability-efficiency trade-off}
\begin{figure}[htbp]
    \centering
    \includegraphics[width=.87\columnwidth]{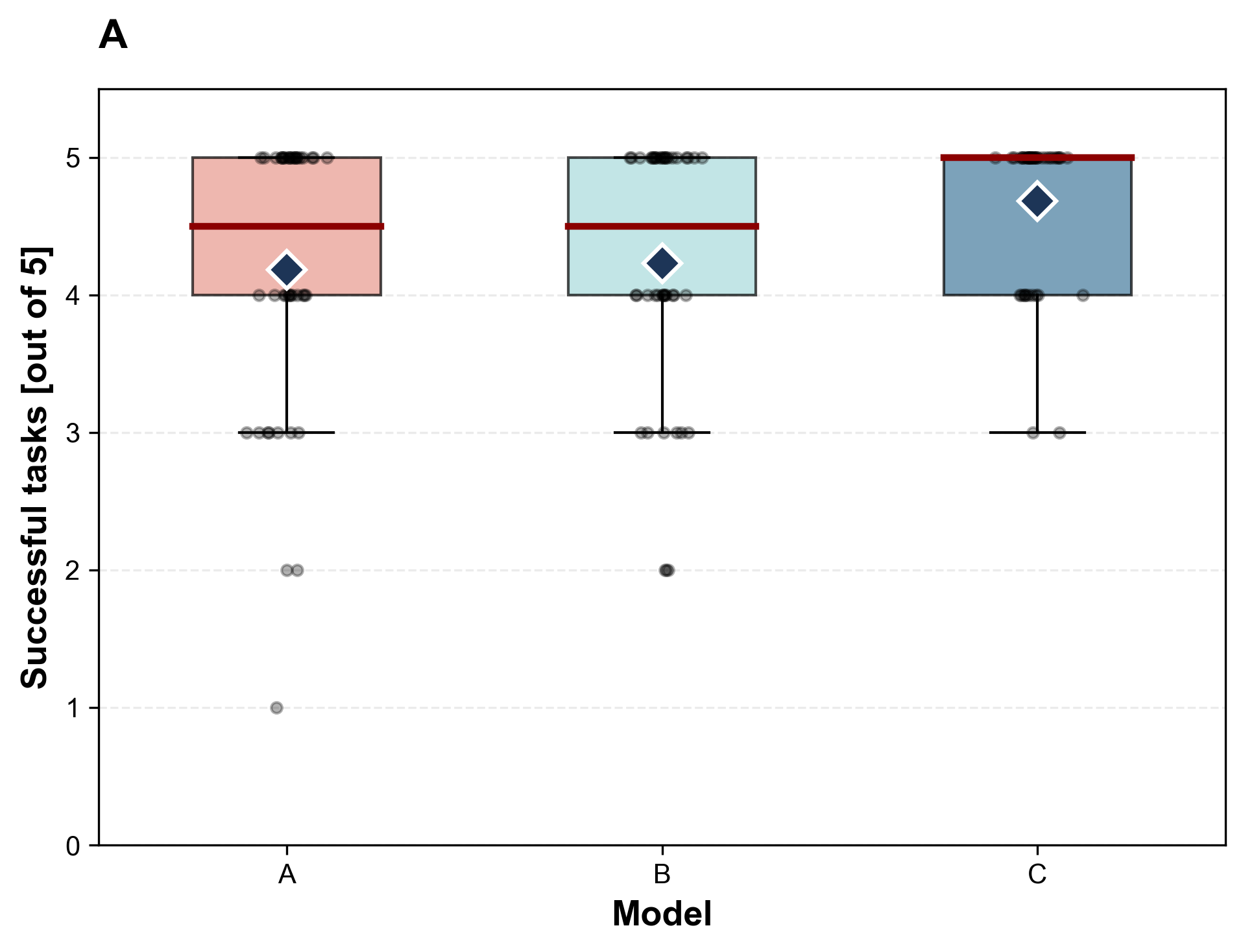}
    \includegraphics[width=.87\columnwidth]{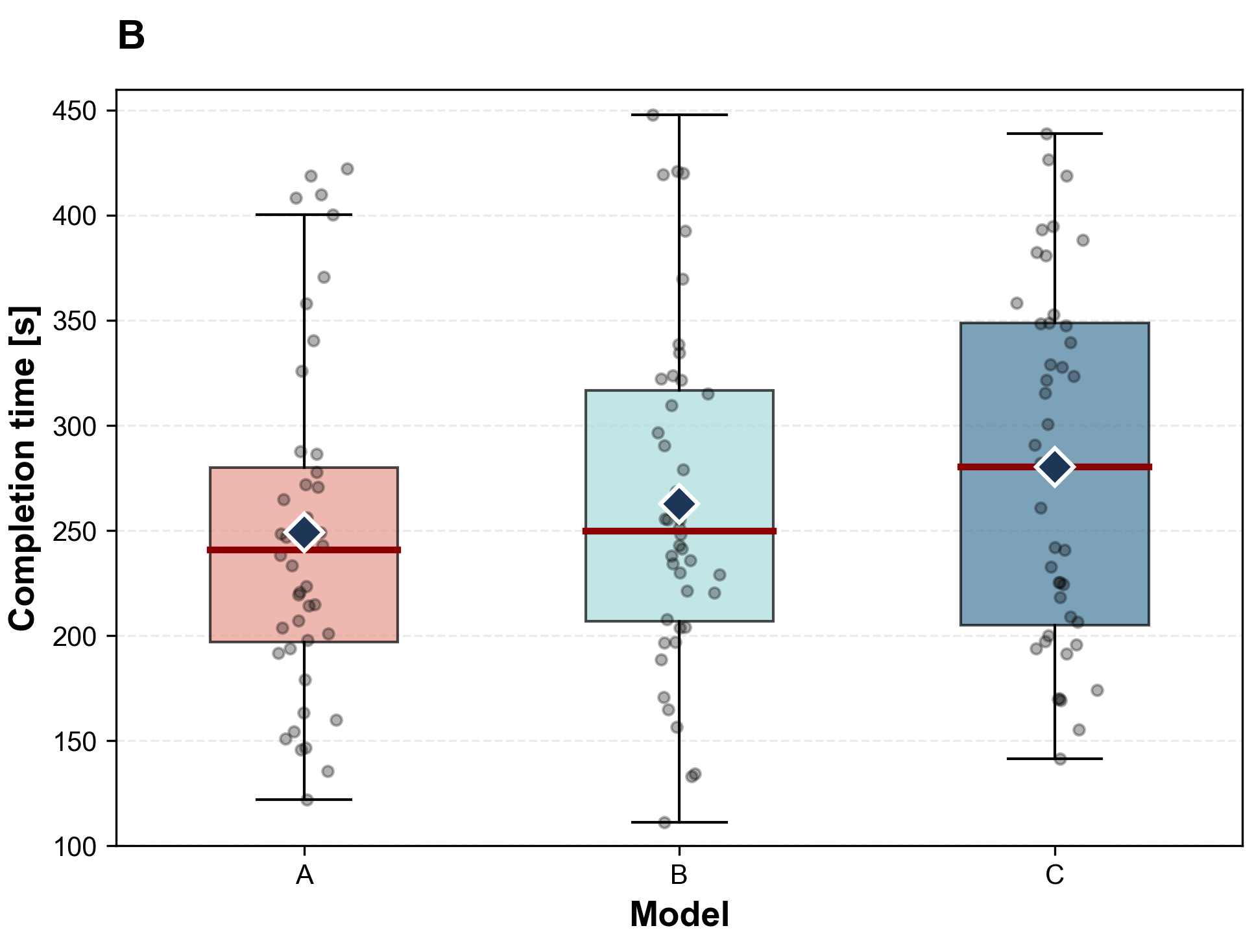}
    \caption{\textbf{Model comparison for success rate and completion time.} (A) Task success rates differed significantly across control models ($n=44$ participants $\times$ 3 trials each, Kruskal-Wallis $H=7.87$, $p=0.020$). Box plots show median (thick red line), interquartile range (box), and range (whiskers). Individual data points (gray circles) show all trial performances. Mean values (navy diamonds) demonstrate Model C achieved highest success rate ($93.6\%$) compared to Models A ($83.6\%$) and B ($84.5\%$). (B) Completion times showed no significant difference across models ($H=2.60$, $p=0.273$). Model A was numerically fastest ($249.2$s), Model B intermediate ($262.7$s), and Model C slowest ($280.3$s), though high inter-participant variability prevented statistical significance. Individual points reveal substantial overlap in distributions. Together, panels suggest a possible speed-accuracy trade-off for Model C.}
    \label{fig:model_comparison}
\end{figure}

The three policy model variants showed differential effects on task performance and completion time (Fig.~\ref{fig:model_comparison}). For task success rates, Model C significantly outperformed Models A and B (Kruskal-Wallis $H=7.87$, $p=0.020$). Model C achieved $93.6\%$ success ($M=4.68/5$ tasks, $SD=0.56$), compared to $83.6\%$ for Model A ($M=4.18/5$, $SD=1.02$) and $84.5\%$ for Model B ($M=4.23/5$, $SD=0.94$), representing a $10$ percentage point absolute improvement in reliability. Post-hoc Mann-Whitney U tests confirmed Model C was significantly better than both Model A ($p=0.012$) and Model B ($p=0.014$).

However, completion times did not differ significantly between models (Kruskal-Wallis $H=2.60$, $p=0.273$). Model A was numerically fastest ($M=249.2$s, $SD=81.8$s), followed by Model B ($M=262.7$s, $SD=81.4$s) and Model C ($M=280.3$s, $SD=85.0$s). The $31$-second difference between Models A and C, while not statistically significant due to high inter-participant variability, represents a $12.5\%$ time difference. 

\noindent
\textbf{Interpreting the speed-accuracy trade-off:}
Model C was trained without background augmentation, creating a narrower learned visual distribution compared to Models A (video-augmented backgrounds) and B (image-augmented backgrounds). We hypothesize that this narrower distribution resulted in more conservative grasping behavior: Model C would only initiate grasps when visual observations closely matched training examples, requiring more precise human positioning. This conservative approach likely traded speed for reliability, participants needed to position the hand more carefully (increasing completion time), but grasps were more likely to succeed when executed (increasing task success rate). Conversely, Models A and B's exposure to diverse backgrounds during training created more permissive grasping policies that triggered across varied viewpoints, enabling faster task execution but with reduced precision.

Participants reported moderate ability to distinguish between models (M = 4.10, SD = 1.65; 54.8\% rated $\geq$5), indicating that model differences were perceptible but not dramatically obvious.
Subjective model preferences revealed both recency bias and genuine model effects. When asked which model was "best overall," participants showed strong recency bias: $22$ selected the model used in Trial $3$, $13$ selected Trial $2$, and only $8$ selected Trial $1$. However, due to counterbalancing, this recency effect distributed across all models 
(Trial $3$ contained Model A for $13$ participants, Model B for $15$, and Model C for $16$). When preferences were decoded to actual models, Models A and C each received $16$ votes ($37\%$ each), while Model B received $11$ votes ($26\%$) from the $43$ participants who expressed a preference. The near-equal preference for Models A and C, despite the recency bias, suggests participants genuinely valued both the intuitive control of Model A and the higher success rate of Model C, with individual preferences likely reflecting task priorities.

\noindent
\textbf{Learning effect dominates model differences:}
To contextualize the model effect magnitude, we compared it to the learning effect observed across trials. The learning effect ($23.3\%$ time improvement from Trial $1$ to Trial $3$, $p<0.001$) was nearly twice as large as the model effect ($12.5\%$ time difference between fastest and slowest models, $p=0.273$). This finding indicates that user adaptation to the variable autonomy paradigm contributed more to overall performance than differences in policy implementation, highlighting the importance of learnable, intuitive control schemes in co-embodied systems. Even with suboptimal policy design, users can develop effective strategies through practice.

\subsection*{User experience evaluation}

\begin{figure}[htbp]
    \centering
    \includegraphics[width=.8\columnwidth]{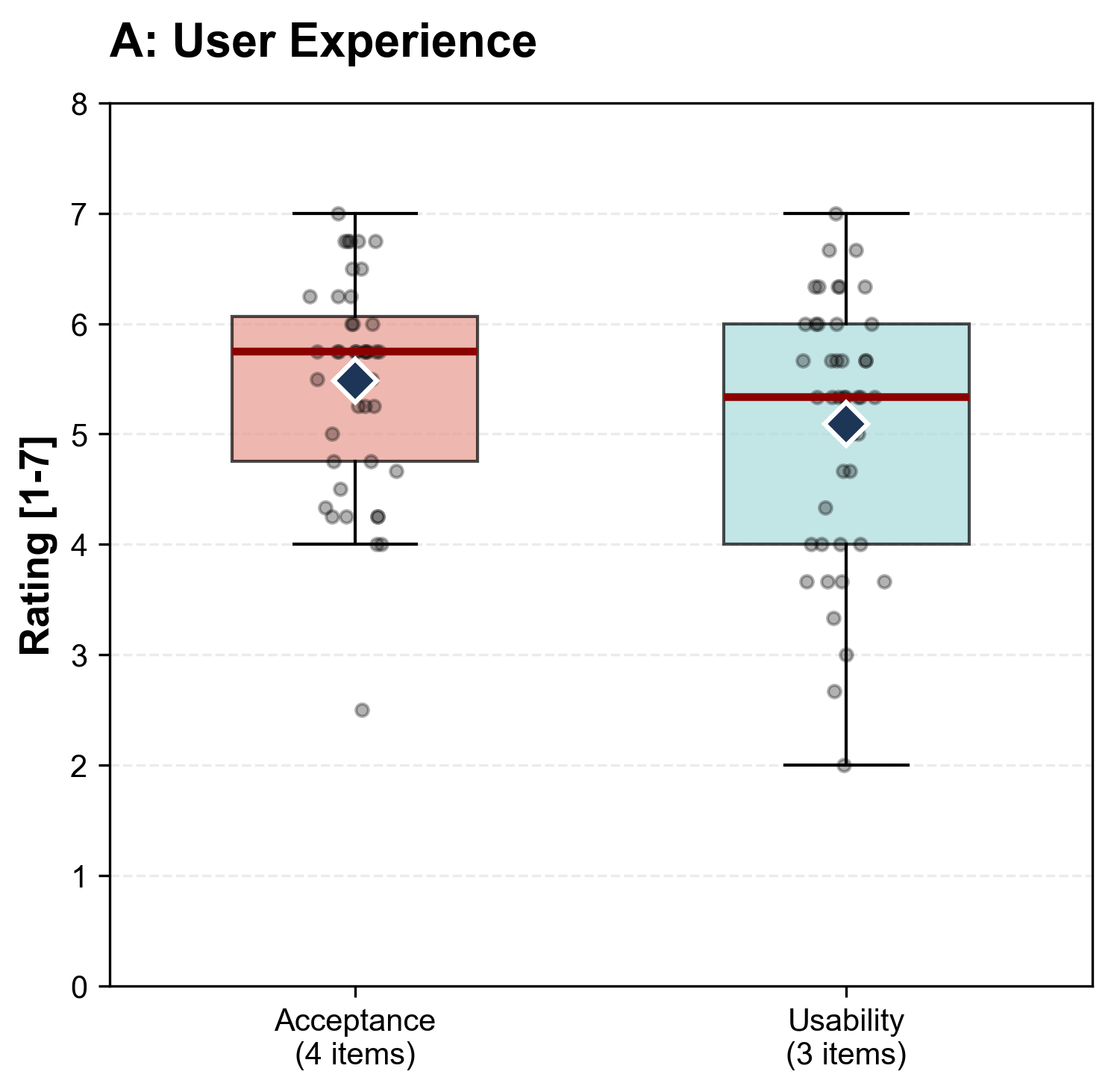}
    \includegraphics[width=.8\columnwidth]{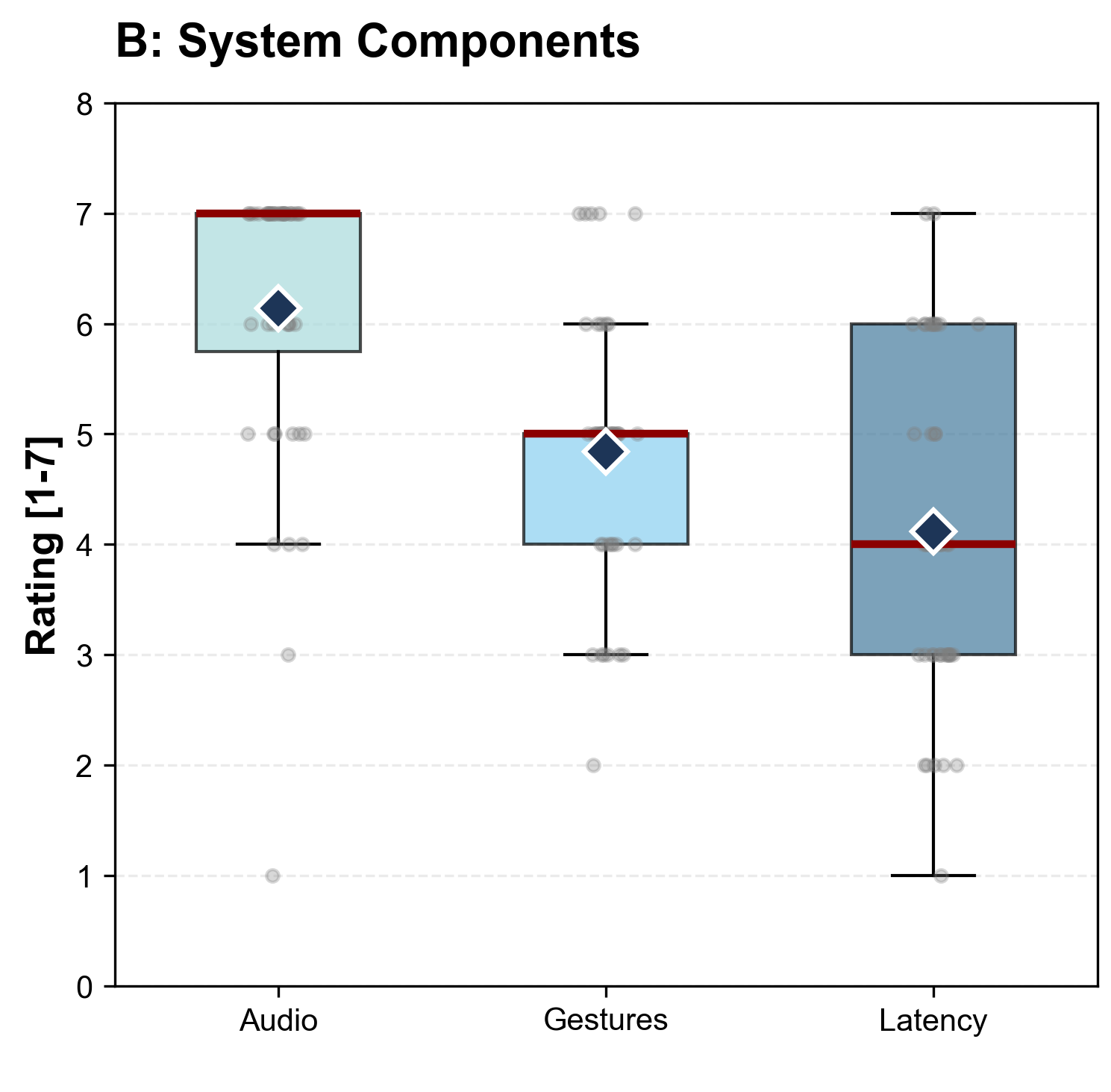}
    \caption{\textbf{UX evaluation of user experiences, system components, and individual measures.} (A) Multi-item scales: Acceptance (4 items, Cronbach's $\alpha=0.709$) and Usability (3 items, $\alpha=0.799$) showed acceptable-to-good internal consistency. (B) System components: Audio feedback received highest ratings, head gestures showed moderate reliability, and latency ratings indicated responsive operation. (C) Other measures: Embodiment ratings were moderate, suggesting users maintained awareness of the robotic hand as a distinct tool rather than a fully integrated body part. Training sufficiency was high, validating the study protocol. Box plots show median (thick red line), quartiles (box), range (whiskers), individual participant scores (gray points), and mean (navy diamond). $n=44$ participants.}
    \label{fig:user_experience}
\end{figure}

UX was evaluated using multi-item scales and individual measures (Fig.~\ref{fig:user_experience}). Acceptance (4 items, Cronbach's $\alpha=0.709$) and Usability (3 items, $\alpha=0.799$) demonstrated acceptable-to-good internal consistency, supporting their use as coherent constructs.

\noindent
\textbf{High acceptance with moderate usability:}
Acceptance averaged $M=5.48$ ($SD=0.96$) on a 7-point Likert scale (Fig.~\ref{fig:user_experience}A), indicating generally positive reception of the co-embodied system. The scale combined ratings of overall system impression ($M=5.70$, $SD=1.19$), willingness to use such a system in daily life if needed ($M=5.52$, $SD=1.21$), perceived helpfulness for bimanual tasks ($M=5.42$, $SD=1.42$), and trust in safe system behavior ($M=5.33$, $SD=1.46$). These high ratings suggest users found value in the collaborative approach and would consider using such technology if necessary. Open-ended responses ($n=37$, 84\% response rate) revealed diverse potential applications, with participants most frequently mentioning food preparation tasks ($n=7$, e.g., ``cutting vegetables, cooking''), writing and creative activities ($n=10$, e.g., ``drawing, painting''), and power tool operation ($n=7$, e.g., ``drilling, assembly work'').

Usability averaged $M=5.09$ ($SD=1.18$) (Fig.~\ref{fig:user_experience}A), the scale encompassed ratings of ease of use ($M=5.11$, $SD=1.24$), sense of control during task performance ($M=5.09$, $SD=1.51$), and comfort and safety during operation ($M=5.07$, $SD=1.42$). While positive, these moderate usability scores suggest room for improvement in system intuitiveness and user control.

\noindent
\textbf{System component performance reveals design priorities:}
Among individual system components (Fig.~\ref{fig:user_experience}B), audio feedback received the highest ratings ($M=6.14$, $SD=1.34$), indicating that explicit state transition announcements (e.g., ``grasped'') were highly valued. This finding underscores the importance of clear communication between the ``two minds'', users need to know precisely when the robot successfully grasps an object.

Head gesture reliability received moderate ratings ($M=4.84$, $SD=1.22$), suggesting functional but imperfect performance. Qualitative feedback from participants reporting discomfort ($n=25$) identified headset tightness as the primary concern ($n=16$ mentions), with participants noting the device ``needed to be quite tight'' for reliable gesture detection. Additional improvement suggestions focused on reducing system weight ($n=12$ mentions), refining gesture control mechanisms ($n=13$, with requests for ``voice input'' or ``EMG-based control''), and enhancing grasp precision for small objects ($n=9$). Future iterations should explore lighter sensor configurations or alternative trigger mechanisms to improve comfort while maintaining hands-free operation.
Furthermore, perceived system latency was relatively low ($M=4.11$, $SD=1.54$, inverted from questionnaire formulation that asked if the user could sense latency where lower scores indicated less noticeable latency), indicating responsive real-time operation that did not significantly impede task performance. 

\noindent
\textbf{Moderate embodiment reflects tool-agent awareness:}
Embodiment ratings revealed that users perceived the robotic hand as a capable tool under their control rather than a fully integrated body part (Fig.~\ref{fig:user_experience}C). Agreement with ``I felt the robotic hand was part of my body during use'' averaged $M=3.86$ ($SD=1.52$), near the scale midpoint. This moderate embodiment may be optimal for co-embodied systems with variable autonomy: complete embodiment might reduce users' awareness of when the autonomous agent is operating or at which autonomy level the system is operating, potentially undermining trust and situational awareness. The ``two minds'' metaphor appears accurate from the user's perspective; users collaborated with a capable partner rather than fully merging with it, allowing them to monitor autonomous performance and retain ultimate veto authority.
Finally, training sufficiency was validated with high ratings ($M=6.23$, $SD=1.24$) (Fig.~\ref{fig:user_experience}C), indicating participants felt well-prepared after the $5-10$ minute practice period, consistent with the rapid learning effects observed in objective performance measures.

\subsection*{Extended practice demonstrates continued learning potential}

\begin{figure}[htbp]
    \centering
    \includegraphics[width=\columnwidth]{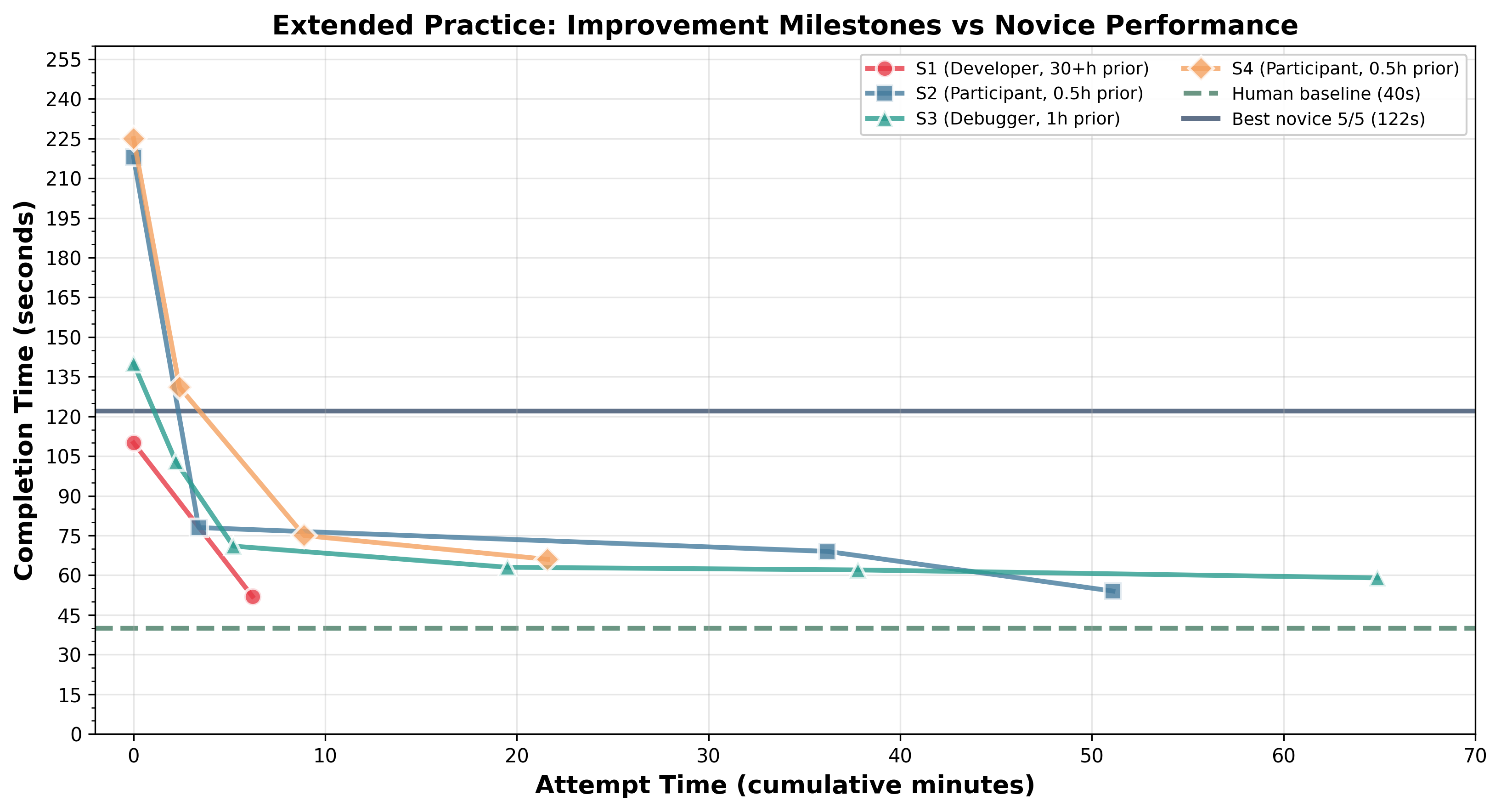}
    \caption{\textbf{Exploratory extended-practice} performance. Improvement milestones are plotted against cumulative attempt time for four extended-practice users in a competitive speedrun format with strategy sharing. The red dotted line indicates the novice Trial 3 mean from the main study, and the green dashed line indicates the human baseline. Because the extended-practice protocol differed from the main study, these comparisons are descriptive rather than controlled. The results suggest substantial performance headroom with additional practice and system familiarity.}
    \label{fig:extended_practice}
\end{figure}

To investigate whether the system supports continued skill development beyond initial exposure, four users with varying prior experience ($0.5-30+$ hours) completed extended practice sessions in a competitive ``speedrun'' format where participants observed each other's attempts and exchanged strategies to minimize completion time. Users repeatedly attempted the five-task battery until achieving personal best performances (Fig.~\ref{fig:extended_practice}).
This protocol was intentionally designed to probe best-case performance and learning potential, not to provide a controlled statistical comparison with the main novice cohort.

In this exploratory extended-practice protocol, final completion times ($M=57.8s$, $SD=6.2s$) were substantially lower than the novice Trial 3 mean ($M=234.8s$, $SD=71.5s$), however, this comparison is descriptive because the extended-practice protocol differed from the main user study (Fig.~\ref{fig:extended_practice}). Notably, expert performance was $53\%$ faster than the best novice $5/5$ performance ($122.0s$), suggesting substantial performance headroom beyond the short familiarization period used in the main study. The fastest expert performance ($52s$) was achieved by the system developer with $30+$ hours of prior experience, suggesting that deep system familiarity accelerates skill acquisition. Even extended-practice users' first attempts ($M=173.2s$) were $26\%$ faster than novice Trial $3$ mean, indicating that modest additional exposure beyond the $30-45$ minute user-study period yields substantial benefits. Users achieved their best performances after $6-65$ minutes of cumulative attempt time, with the best five performances averaging $58.0s$ ($SD=4.8s$), approaching but still not quite matching the human baseline of $40s$. These exploratory results suggest that habitual use may yield further efficiency gains, but longitudinal studies with target users are needed to quantify this effect under realistic daily-use conditions.

\section*{Discussion}

\textbf{Co-embodiment is a viable approach for assistive robotics.}
We introduced \emph{co-embodiment with variable autonomy}, a collaboration paradigm in which a human and an autonomous policy share a single physical body and operate at different autonomy levels across task phases. In contrast to shared autonomy approaches that continuously blend or arbitrate commands, our system is designed around a task-phase decomposition: the user provides intent, timing, and gross positioning of the wearable hand, whereas the policy provides reliable grasp execution once the hand is placed near a target object. Importantly, the two ``minds'' operate with mutual autonomy during search and grasping phases: the user moves the arm while the policy continuously evaluates sensory input and initiates a grasp when conditions are met, both agents acting autonomously but coupled through the shared physical body. After grasp acquisition, the user proceeds with tool actuation and task timing while retaining a release override at all times. Two ``minds'' working autonomously but remaining implicitly coupled through the shared physical substrate are able to complete complex bimanual tasks with fast learning adaptation and further improvement given prolonged system interaction, which demonstrates co-embodiment as a viable approach for assistive robotics.

\textbf{Rapid learning reflects mastery of transitions and timing.}
Across the five bimanual tasks, completion time improved quickly over the first trials and then stabilized, suggesting that participants primarily learned \emph{when} and \emph{how} to position the hand to enable successful autonomous closure, rather than learning the task itself. This pattern indicates that effective use of co-embodied assistance depends strongly on a user forming a reliable mental model of \emph{(i)} what constitutes a ``good'' pre-grasp hand pose and approach trajectory and \emph{(ii)} when to pause, reposition, or proceed to actuation. The short learning curve supports the practicality of phase-based responsibility sharing for everyday manipulation.

\textbf{Reliability-efficiency trade-offs emphasize engagement design.}
Comparing policy variants revealed a trade-off between success and speed: more reliable grasping can come at the cost of increased time, likely because users must position the hand more carefully to satisfy the policy's implicit conditions for closure. These results highlight that co-embodied assistance with variable autonomy is governed by two coupled problems: (1) \emph{engagement:} how predictable, stable, and learnable the autonomy transitions and trigger conditions are under user-driven motion and (2) \emph{execution:} how robustly the policy closes and maintains a grasp once engaged. Improving open-world usefulness, therefore, requires not only more generalizable grasp policies but also engagement mechanisms that are legible to users (e.g., confidence cues that indicate when a grasp is likely to succeed).

\textbf{Embodiment and agency under co-embodied variable autonomy.}
Participants reported moderate embodiment and high acceptance, suggesting that physical integration and the two mind paradigm support intuitive collaboration without requiring continuous attention to low-level control. In co-embodied settings, the user's arm motion shapes contact conditions even when finger motion is autonomous, yielding a form of implicit collaboration that differs from both teleoperation and independent autonomy. This interaction structure may be especially valuable in bimanual tasks, where the user can focus on high-level timing and actuation while the robotic hand contributes precise, repeatable grasp acquisition.

\textbf{Practice indicates headroom beyond novice performance.}
Extended practice data suggest that performance continues to improve beyond the rapid novice plateau. This supports the view that co-embodied assistance is a skill that users can refine through experience, learning object-specific approach heuristics and more consistent engagement strategies. Such learning effects imply that short lab sessions may underestimate real-world efficiency gains for habitual users. Furthermore, it highlights that even with suboptimal policies, the human mind is able to successfully collaborate with the second mind of the hand without explicit control signals.

\subsection*{Limitations}

\textbf{Able-bodied participants with two hands.}
All participants in our main study were able-bodied and had two functional biological hands. This is a critical limitation: the study demonstrates feasibility, learnability, and acceptance of the co-embodied control paradigm, but it does not directly quantify functional independence gains for users with upper-limb impairment. Future work must evaluate the system with participants who have relevant impairments and in ecologically valid settings.

\textbf{Short-term lab exposure and constrained task set.}
The protocol captures rapid familiarization rather than long-term comfort, fatigue, and trust calibration over sustained daily use. Moreover, our task and object set was limited, and although we tested in a non-sterile environment, broader generalization to clutter, lighting variation, and novel objects remains an open challenge.

\textbf{Hardware and intent interface constraints.}
The wearable form factor and hands-free triggering provide a low-effort interaction channel, but ergonomics and interface reliability will shape real-world adoption. Participants' feedback on comfort and control modalities motivates exploring alternative intent channels (e.g., voice or EMG) and improving wearable design for prolonged use.

\textbf{Outlook.}
Future work should \emph{(i)} validate assistive benefits in target user groups, \emph{(ii)} expand robustness to open-world object variation, and \emph{(iii)} improve engagement legibility through explicit confidence cues and user-guidance signals. More broadly, co-embodiment with variable autonomy offers a promising direction for assistive manipulation, enabling collaboration through physical integration and the two minds paradigm, while avoiding continuous command blending.

\section*{Materials and Methods}

\subsection*{Co-embodied robotic hand system}

\begin{figure}[htbp]
    \centering
    \includegraphics[width=\columnwidth]{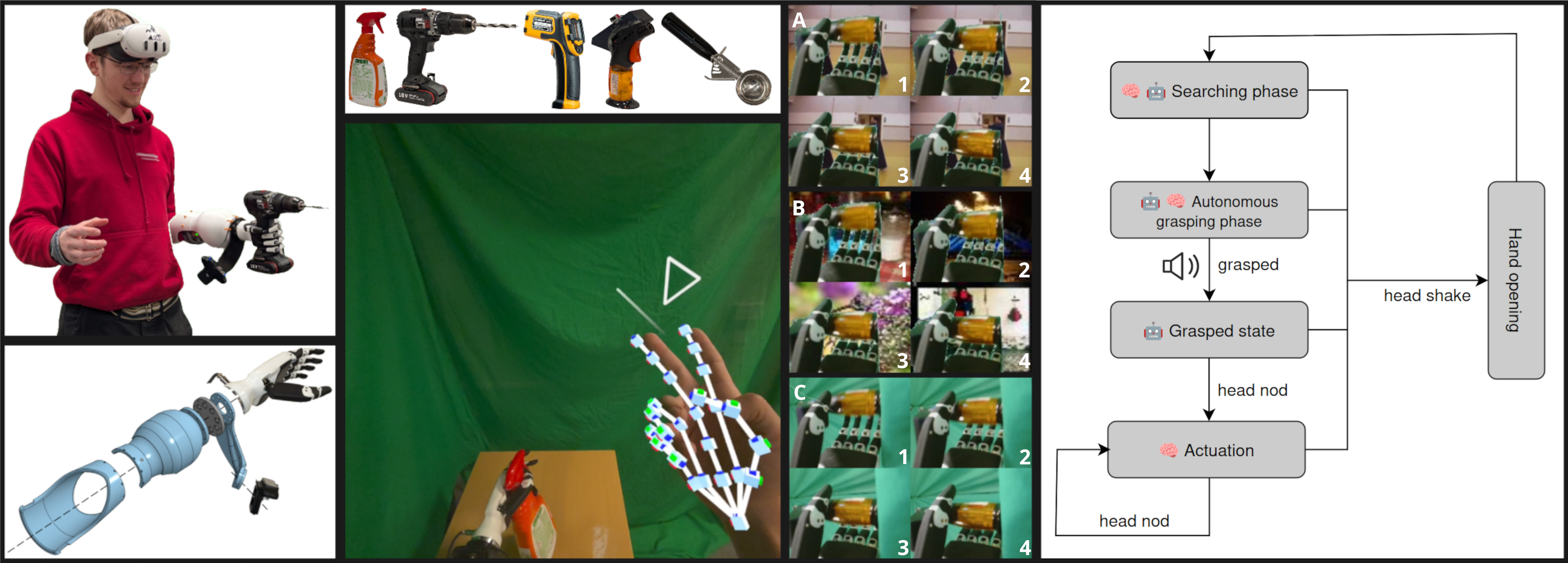}
    \caption{Overview of the system: Hardware setup, Objects, Teleoperating POV, Example dataset frames, Control loop flowchart}
    \label{fig:system}
\end{figure}

\subsubsection*{Hardware configuration}

We developed a wearable robotic hand system that enables variable autonomy between human and robot within a shared physical body. A robotic hand, 6 DoF RH56DFX-2L from Inspire-Robots, is mounted on the user's left forearm via a custom 3D-printed structure with a USB camera (Logitech C922 PRO HD) and secured with adjustable straps. Users wear an AR headset (Meta Quest 3) positioned on the forehead, not worn covering the eyes, for audio feedback and hands-free gesture recognition through head motions (Fig.~\ref{fig:system}). The system is connected to a stationary PC equipped with a RTX 4090 for model inferences.

\subsubsection*{Training the Robotic Hands "mind"}

We collected $50$ demonstrations for each object (spray bottle, power drill, infrared thermometer, lighter, and ice cream scoop) in a green screen setup (Fig.~\ref{fig:system} middle) by teleoperating the left-handed robotic hand with the user's right hand. This was achieved by adjusting the teleoperation setup described in~\cite{koczy2025learning} so it would remap commands form the user's right hand onto the left-handed robotic hand.

We post-processed the total of $250$ demos and manually added, based on the video and force recordings, a \textit{grasped} signal to each demo, a clear indicator of when the object is securely grasped. This signal was set to $0$ from the beginning of the demo until the object was securely grasped from which it was set to $1$. Furthermore, the information which finger is used for the particular object was also embedded into each demo by setting the respective signal of that finger (index, index and middle, or thumb) to $1$ at the same time as the \textit{grasped} signal and the other two signals to $0$. 
In the next post-processing step we employed Segment Anything Model 2 (SAM 2) \cite{ravi2024sam2segmentimages} to replace the green-screen background for Model A with randomly selected videos from the HMDB51 Video Database \cite{6126543}. This forces the model to be more robust against changes in the background while still preserving the temporal consistency. For Model B, we instead substituted the greenscreen with random frames drawn from ImageNet\cite{5206848}, this variant also effectively randomizes the background for the model, but does not preserve any notion of temporal consistency in the background. Finally Model C demo data left the background unchanged.

The post-processed expert demos are used to train the three model variants using visuomotor policies\cite{chi2024diffusionpolicyvisuomotorpolicy}.
We set the maximum training to $600$ epochs, enabling early stopping with a patience of $100$ epochs using a $10\%$ held out validation set.
The Models are trained to output a delta joint angle for each joint to the hand controller, as well as the \textit{grasped} signal and the three signals for which finger to use for activation. Once the \textit{grasped} signal is above the predefined threshold of $0.8$ the highest activation of the index, index \& middle, or thumb signal is taken $0.5$ seconds after and used to determine which finger to use for activation in the next phase.

\subsubsection*{Co-embodiment with variable autonomy level paradigm}

In a co-embodied setup, commands are not blended together as in shared autonomy, nor is control relinquished fully to one or the other partner; our system has "two minds" in one body. This, however, does not mean that there is no communication or feedback mechanism. In detail, we can classify this particular task at hand into four phases and detail the level of autonomy and task responsibility each "mind" has to exemplify the co-embodiment notion.
Note that the paradigm of co-embodiment with variable autonomy level is not restricted to these particular instances shown.

\textbf{i) Searching phase:} Both minds are fully autonomous. The human freely positions the robotic hand in 3D space through natural arm movements. A visuomotor diffusion policy runs continuously in the background, monitoring visual input for known objects but remaining passive until an object enters grasping range. The human has a simple higher-level control option, namely opening the hand via a head-shaking 
gesture at any time, overwriting whatever the policy predicts. Once the hand is fully open, however, the policy is free to act autonomously again. 

\textbf{ii) Autonomous grasping phase:} Both minds are fully autonomous. When the hand approaches a target object, the policy automatically executes a learned grasp trajectory appropriate for that object type. Critically, the human can continue moving the arm during this phase; the policy adapts its finger commands in real-time based on changing visual observations, creating implicit physical coupling between the "two minds" even during autonomous operation. The searching and this phase creates a unique form of collaboration of the two "minds", where the positioning of the human directly influences the quality of the grasp policy, and its likelihood of success, while the human is also observing the robot's grasping effort and intuitively adjusts to help the overall task success. This bidirectional physical coupling, where both "minds" have full autonomy but neither agent has complete control bit both influence the outcome, is a hallmark of the co-embodiment setting.

\textbf{iii) Grasped state:} Human mind in full control, robot secures the grasp. Once the policy determines a successful grasp, it tightens the grasp configuration and signals successful completion via audio confirmation ("grasped") through the AR headset, and sets which finger will be used for actuation. Control authority is now fully with the human, who can now position the grasped object and trigger its actuation.

\textbf{iv) Human actuation:} Human in full control, robot waiting for trigger gesture. The user performs task-specific actions by executing a nodding head gesture to activate the object (e.g., squeezing a spray bottle trigger, powering a drill). The same nodding gesture deactivates the object. The human retains full spatial control through arm movement while the robot maintains the stable grasp. Once the human is satisfied with the task or wants to preempt it, a head-shaking gesture opens the hand and returns the system to phase one.

\subsection*{Task battery and experimental environment}

\subsubsection*{Experimental workspace}

The experimental workspace was placed in a busy live laboratory environment where other unrelated work and conversations occurred (Fig.~\ref{fig:setup}). This setup was chosen to increase ecological validity and introduce natural background variation that would test policy robustness to real-world conditions. The five tools were arranged on a central table from which participants grasped them, with 
dedicated work areas for each task positioned clockwise around the workspace. Changing lab conditions can be seen best via the supplementary videos.

\subsubsection*{Task descriptions}

Participants performed five bimanual tasks requiring coordination between their biological right hand and the co-embodied robotic left hand:

\textbf{Ice cream scooping:} The ice cream container and corresponding bowl were positioned on a table. The participant used their right (human) hand to hold the ice cream container in place while the robotic hand (holding the scoop) was used to scoop ice cream. The triggering of the ice cream scoop was then used to deposit it into the bowl held/fixated by the participant.

\textbf{Power drilling:} An elevated table contained a wooden board that had to be held against a fixture with the right hand. The task was to grasp the drill with the robotic hand, trigger it on, and drill a hole through the wooden board.

\textbf{Spray-and-wipe:} Spraying and wiping was done on a designated area marked on a whiteboard. The participant had to successfully spray onto the marked area using the robotic hand, then wipe the board clean with a cloth using the right hand.

\textbf{Infrared thermometer:} A common kitchen water boiler and a cup were placed on a table. The participant filled the cup with water using the right hand, and used the robotic hand to grasp the thermometer and trigger a temperature measurement of the water.

\textbf{Lighter:} Two candles were placed in candle holders on a table. Before triggering the lighter, the participant had to disengage the safety mechanism with their right hand while the robotic hand held the lighter, then light both candles.

For safty and usability the Drill, Lighter, and Ice cream scoop recived slight modifiaction outlined in Supplamentary section S4.

\subsection*{User study}
The user study objectives were to quantify: \emph{i)} user adaptation and system learnability using objective task performance measures (success rate + completion time). \emph{ii)} the effect of the different models (A, B, C),  \emph{iii)} user-reported scoring of, user experiences (Acceptance, usability), System components (audio, gestures, latency), and embodiment and training time, and \emph{iv)} If prolonged exposure leads to further performance gains.
In total $n=44$ participants were recruited from the authors’ local university $35$ male, $8$ female, $1$ prefer not
to say; age $M = 28.27$ years, $SD = 6.24$, range $17-53$), and all completed the full experiment ($30–45$ min per participant).
Participants reported diverse robotics experience ($7$ none, $12$ beginner, $13$ intermediate, $12$ advanced) and minimal prosthetics/wearable robotics experience ($39$ none, $5$ beginner). The majority were right-handed ($37$ right, $7$ left) and all wore the robotic hand on their left arm. The user study was conducted in full compliance with the ethical regulations and institutional review procedures of the institute.

\subsubsection*{Study design and conditions}
Participants were initially provided with clear information about the study aims,
procedures, potential risks, and data handling process , and they were requested to give informed consent before taking part.
Participation was entirely voluntary, and participants could withdraw at any time during the study without reasons.

Before the formal study, participants received standardized instructions and watched a demonstration video (Supplementary Movie M2), then completed a $5-10$ minute hands-on practice session to familiarize themselvesf with the system. 
A safety briefing covering proper handling of the power drill and lighter was included in this phase.
During experimental trials, participants attempted the complete five-task setup within a $7$-minute trial limit. Task order within each trial was not constrained, allowing participants to adopt personalized strategies; most followed the sequence shown in the demonstration video. An incentive structure was presented where participants would receive an additional $15$ SEK for each trial in which they successfully completed all five tasks (5/5), with emphasis that completion time was not critical unless the participant personally chose to optimize speed.


We used a within-subject repeated-measures design principle in which each participant completed three trials to capture short-term learning effects. To dissociate user adaptation from autonomous policy robustness, each trial was conducted with a different policy model (A, B, or C). Model order was assigned using Latin-square counterbalancing across the three trials to mitigate the order effect (ABC: $n=8$, ACB: $n=8$, BAC: $n=8$, BCA: $n=7$, CAB: $n=7$, CBA: $n=6$).

After completing all trials, participants completed questionnaires assessing user experience, system components and other measures on $7$-point Likert scales, ranked model preferences, and provided open-ended feedback. (see S1 for more details about the questionnaires and aualitative feedback).

\subsubsection*{Task success criteria}

A single trained experimenter scored each task via the video recording as success $(1)$ or failure $(0)$ using predefined operational criteria. Success required: \emph{(1)} successful autonomous grasp of the target object, \emph{(2)} stable maintenance of grasp during positioning, and \emph{(3)} successful actuation of the tool at the designated target location.
Tasks were marked as failed if: the object was dropped outside the tool table before task was completed.
The $7$-minute time limit per trial included time for recovery from minor errors. 
Video recordings of all sessions were retained for verification, videos from participant that gave media consent are displayed on the project website, see supplementary S2. 

\subsubsection*{Extended practice protocol}

To assess learning beyond novice exposure, four users with varying prior experience with the system ($0.5$, $0.5$, $1$, and $30+$ hours + at least $5$h+ of observation of other participants performing the task) completed extended practice sessions in a competitive "speedrun" format. Participants observed each other's attempts and openly exchanged strategies to minimize completion time. Users repeatedly attempted the five-task battery, with each attempt timed and recorded. Practice continued until participants felt they had achieved their personal best  performance or showed no further improvement over multiple consecutive attempts.
We tracked improvement milestones (new personal best times) against cumulative attempt time to visualize skill acquisition trajectories.
As model A was the \textit{fastest} Model during the user study all extended-practice users used model A. Furthermore, they could \textit{reset} at any time, meaning if an initial task would fail, the setup was reset immediately, reducing time on unpromising attempts. Furthermore, an additional constraint was added for the experts: the tool had to be returned to the tool tables and not dropped (as novice participants were allowed).

\subsection*{Statistical analysis}

\subsubsection*{Learning effects}

We assessed trial-to-trial changes in completion time and task success using non-parametric repeated-measures tests. Friedman tests were used to test for overall effects of trial number across the three trials, followed by Wilcoxon signed-rank tests for pairwise comparisons where the overall effect was significant. For completion time, we computed Cohen's $d$ effect size for the Trial $1$ vs Trial $3$ comparison. A sensitivity analysis restricting to participants with non-decreasing 
task success across trials confirmed the robustness of learning effects (Supplementary Note S3).

\subsubsection*{Model comparison}

We tested for differences in task success and completion time across the three model variants (A, B, C) using Kruskal-Wallis tests. Where the omnibus test was significant, we conducted planned post-hoc comparisons using Mann-Whitney $U$ tests. For task success, we compared the best-performing model (Model C) against Models A and B. No correction for multiple comparisons was applied as these were planned comparisons following a significant Kruskal-Wallis test.

\subsubsection*{Task-specific analysis}

Individual task success rates were analyzed descriptively across trials. Given the small number of trials per task ($3$) and ceiling effects for most tasks, formal statistical tests were not conducted for individual task trajectories.

\subsubsection*{Questionnaire analysis}

Internal consistency of multi-item scales (Acceptance, Usability) was assessed using Cronbach's $\alpha$. Values $\geq 0.70$ were considered acceptable. Individual items and scale scores were analyzed descriptively (mean, standard deviation, median, range). Open-ended responses were analyzed using content analysis to identify recurring themes, with frequency counts reported for major categories.
All statistical tests used a significance threshold of $\alpha = 0.05$ (two-tailed). 
Sample size ($n=44$) provides $>80\%$ power to detect a medium effect size (Cohen's $d=0.5$) for within-subjects repeated-measures comparisons at $\alpha=0.05$, based on G*Power 3.1 calculation~\cite{faul2007g} assuming three measurement time points and moderate correlation ($r=0.5$) between measurements.




\section*{APPENDIX}
\subsection*{Additional questionnaire analysis S1}

Participants completed post-experiment questionnaires assessing multiple dimensions of UX using $7$-point Likert scales (1 = strongly disagree/very poor, 7 = strongly agree/very good).

\textbf{Acceptance} was measured using four items: overall system impression, willingness to use such a system in daily life if needed, perceived helpfulness for bimanual tasks, and trust in safe system behavior. Internal consistency was acceptable (Cronbach's $\alpha = 0.709$). The acceptance score was computed as the mean across the four items.

\textbf{Usability} was measured using three items: ease of use, sense of control during task performance, and comfort and safety during operation. Internal consistency was good (Cronbach's $\alpha = 0.799$). The usability score was computed as the mean across the three items.

\textbf{Individual system components} were rated separately: audio feedback quality, head gesture reliability, and perceived system latency. For latency, the question was phrased as "I could sense delay/latency in the system" and scores were inverted 
for analysis such that higher values indicated lower perceived latency.

\textbf{Embodiment} was assessed using a single item: "I felt the robotic hand was part of my body during use" 

\textbf{Training sufficiency} was assessed with: "The practice period was sufficient to learn how to use the system" 

\textbf{Model distinguishability and preference:} Participants rated their ability to distinguish between the three models (1 = could not tell them apart, 7 = very clearly different) and indicated which model they felt was "best overall."

\textbf{Open-ended questions} asked participants to: \emph{(i)} describe tasks they would like to perform with such a system, \emph{(ii)} suggest improvements, and \emph{(iii)} describe any discomfort experienced. 

\begin{figure}[htbp]
    \centering
    \includegraphics[width=\columnwidth]{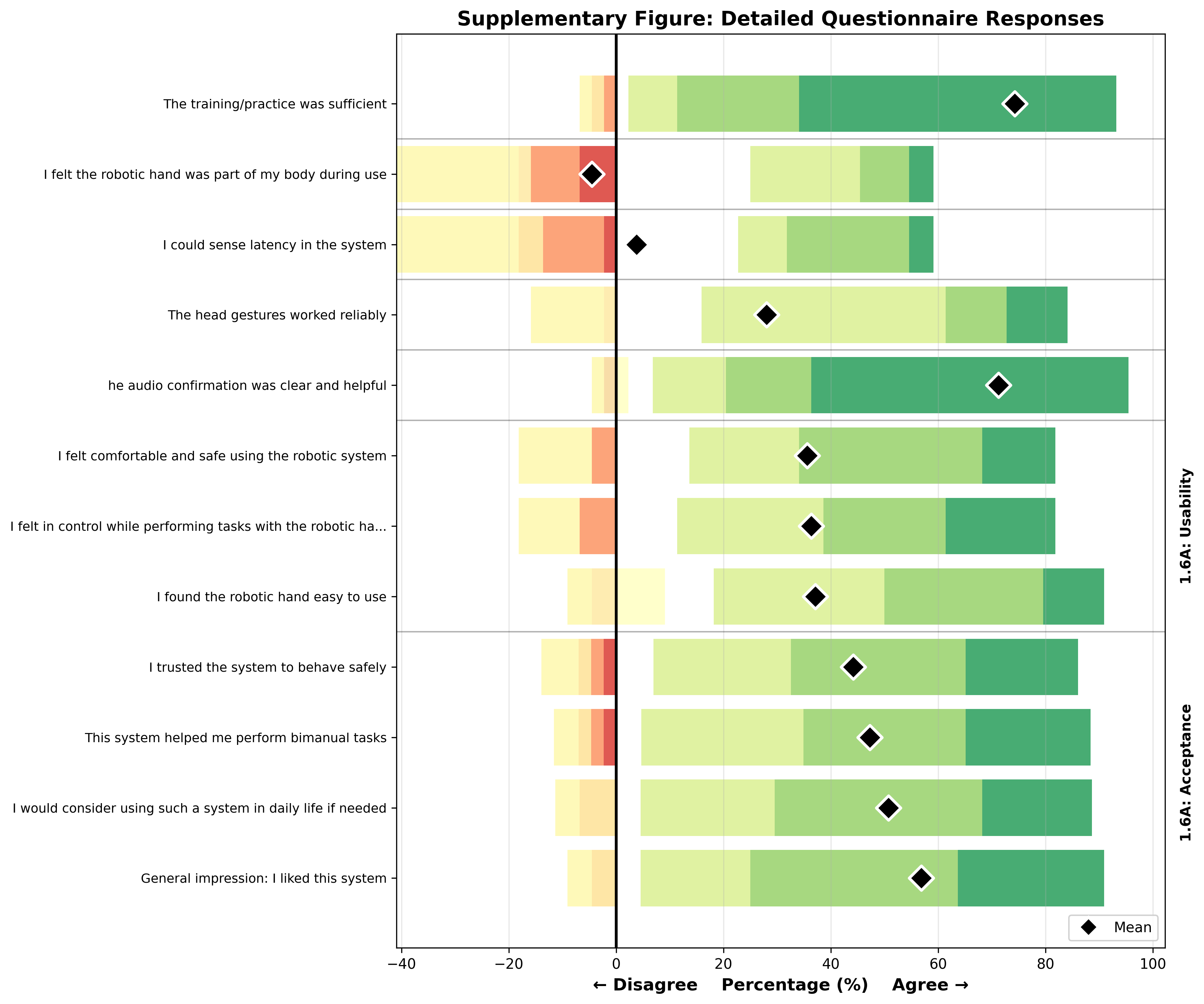}
    \caption{Likert responses for all individual items on the questionnaire}
    \label{fig:S_likert}
\end{figure}

A detailed overview of all liker score responses is shown in Fig.~\ref{fig:S_likert}.

\subsection*{Project website content S2}

The project website can be found at \texttt{https://co-embodiment.github.io/} \cite{website} and contains:
\begin{itemize}
    \item Video recordings of user study trials
    \item Demonstration video shown at the beginning of user study
    \item 3D models of hand mount parts
\end{itemize}
\subsection*{Robustness of learning effects S3}

To verify that improvements in completion time reflected genuine efficiency gains, rather than participants ``failing faster'' by abandoning tasks, we conducted a sensitivity analysis restricted to participants who maintained or improved task success across all three trials (\(n=26\) of \(44\), \(59.1\%\)).

Results remained consistent with the primary analysis:
\begin{itemize}
  \item Completion times: \(M=286.9\,\mathrm{s}\) (Trial 1) \(\rightarrow\) \(217.7\,\mathrm{s}\) (Trial 3)
  \item Improvement: \(24.1\%\) (vs.\ \(23.3\%\) in the full sample)
  \item Friedman test: \(\chi^{2}=9.54\), \(p=0.0085\) (vs.\ \(\chi^{2}=15.95\), \(p<0.001\) in the full sample)
  \item Cohen's \(d\): \(0.94\) (identical to the full sample)
  \item Post-hoc tests: T1 vs.\ T2 (\(p=0.013\)), T1 vs.\ T3 (\(p<0.001\)), T2 vs.\ T3 (\(p=0.208\))
\end{itemize}

\begin{figure}[htbp]
    \centering
    \includegraphics[width=.8\columnwidth]{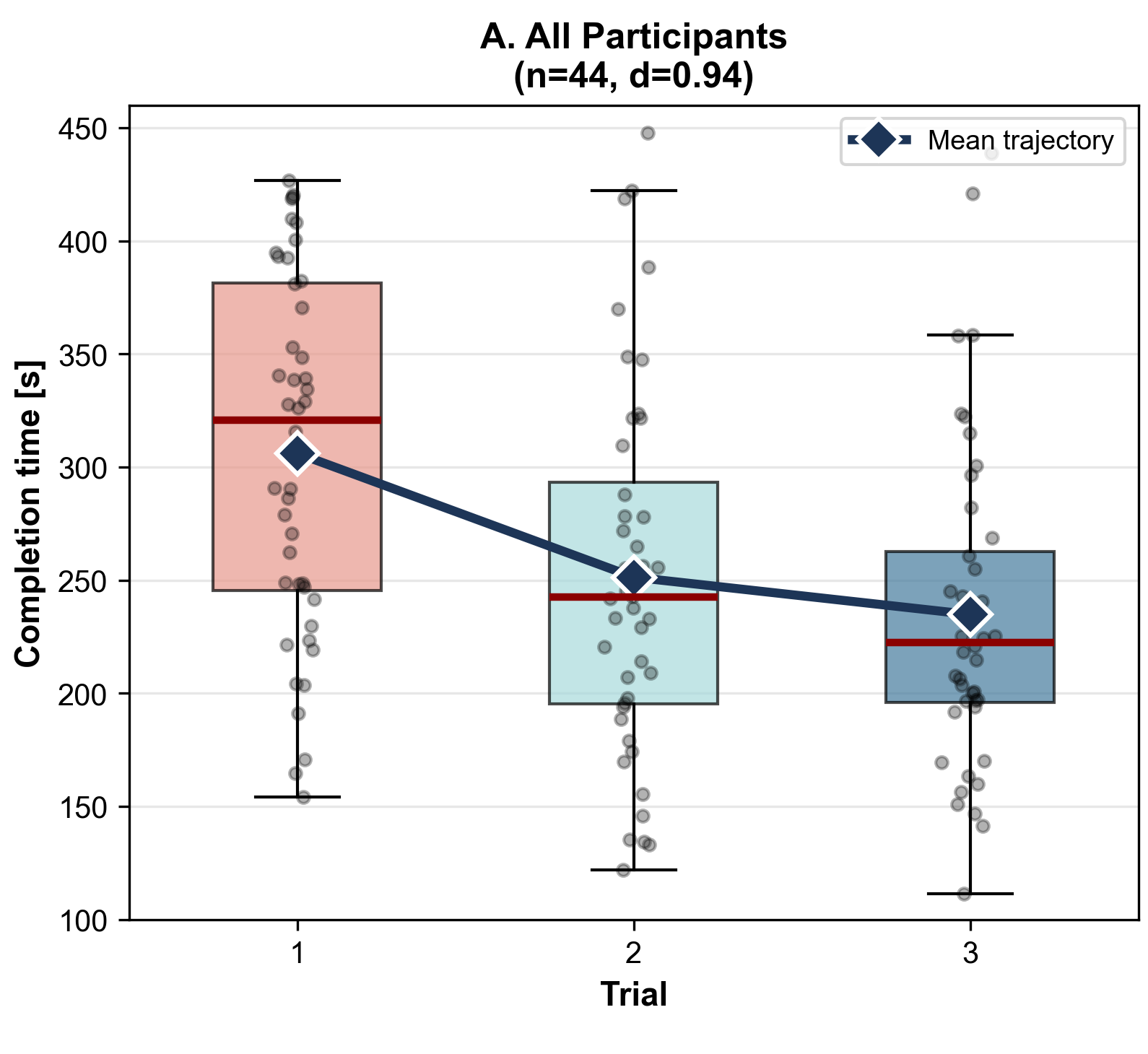}
    \includegraphics[width=.8\columnwidth]{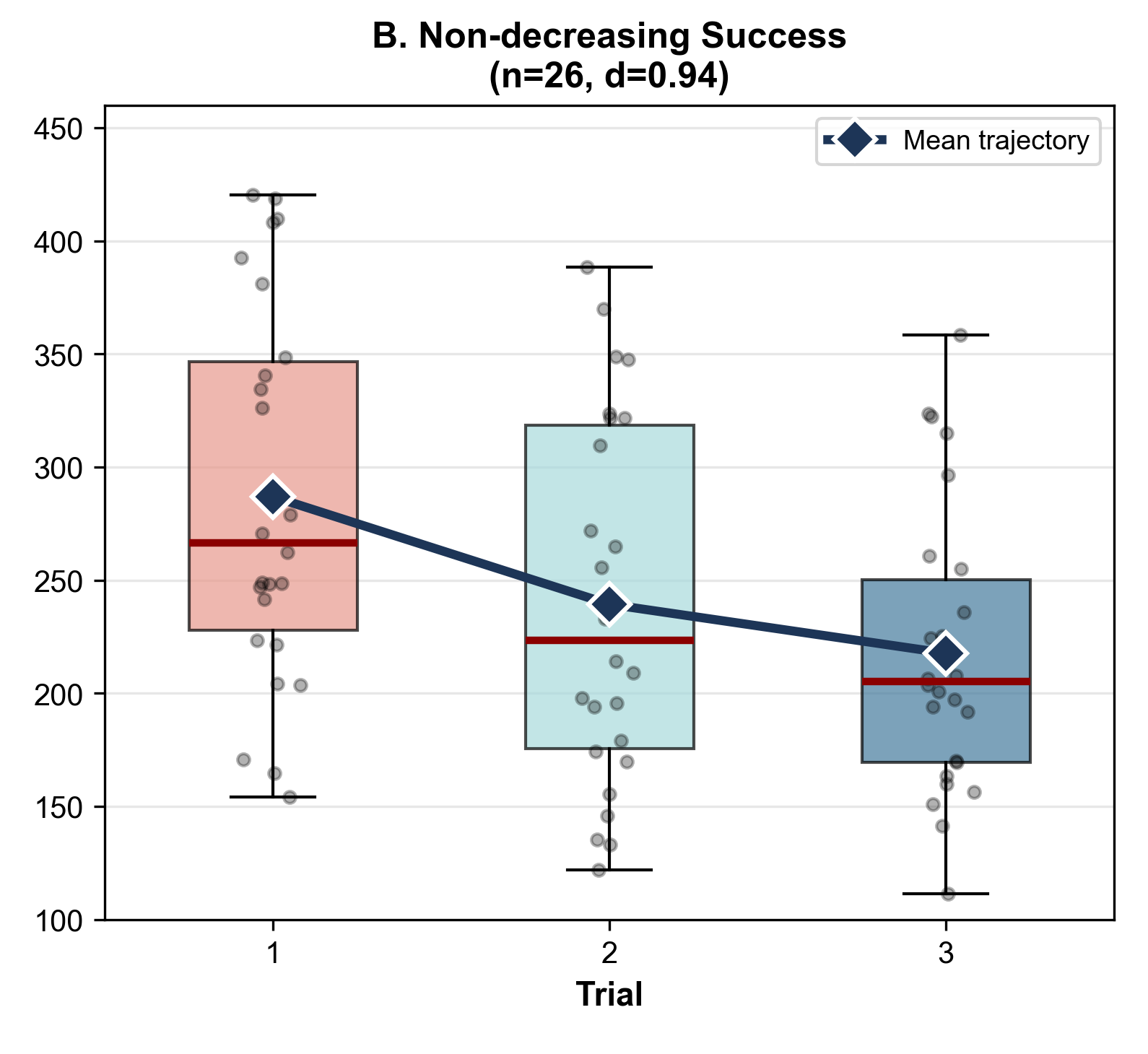}
    \includegraphics[width=.8\columnwidth]{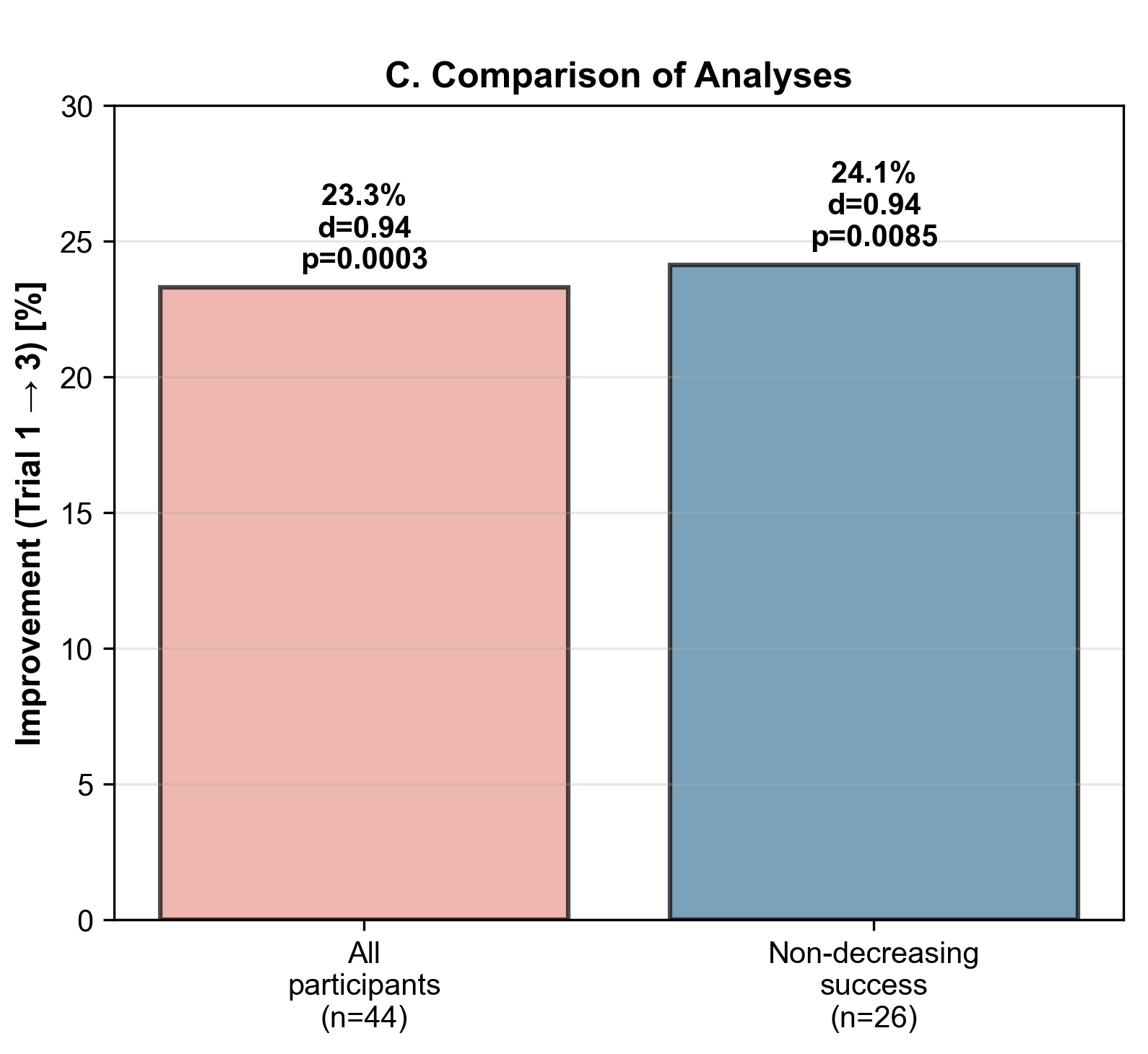}
    \caption{\textbf{Robustness of learning effects.}
Sensitivity analysis restricted to participants who maintained or improved task success across all three trials ($n = 26$ of $44$) confirmed that completion-time improvements were not driven by participants ``failing faster.'' Completion time decreased from $M = 286.9$~s in Trial~1 to $M = 217.7$~s in Trial~3, corresponding to a $24.1\%$ improvement, closely matching the $23.3\%$ improvement observed in the full cohort. The learning effect remained statistically significant (Friedman test: $\chi^2 = 9.54$, $p = 0.0085$), with the same Trial~1 vs.\ Trial~3 effect size as in the full sample (Cohen's $d = 0.94$). These results support the interpretation that participants became more efficient while maintaining task performance.}
    \label{figS:learning_robust}
\end{figure}

Only 6 participants (\(13.6\%\)) exhibited faster completion times accompanied by decreased success (Trial 1 to Trial 3), with time reductions ranging from \(35\) to \(188\,\mathrm{s}\). The identical effect size (\(d=0.94\)) across both analyses confirms that learning effects were not driven by premature task abandonment. The slight reduction in statistical power with filtered data (\(p=0.0085\) vs.\ \(p<0.001\)) is expected given the reduced sample size but does not alter the conclusion of significant learning (Fig.~\ref{figS:learning_robust}).

\subsection*{Hardware tool modifications S4}
Not all of our test objects originally allowed for a reliable grasping and activation using our Inspire Hand due to lack of strength in its joints and range of motion. Therefore, we had modified some of them in a way that we find does not make the tasks easier for our policy but solves hardware limitations. Objects limitations were as follows:
\begin{itemize}
    \item Drill - spring in the button was modified to lower the required force to press it. Without it the index finger very often could not press it and turn on the drill.
    \item Lighter - force required to press the button to the point of creating spark was also too large so we replaced the piezoelectric igniter with an electronic one. The spark is controlled by a button hidden inside and the circuitry was placed in the black box on top of the lighter.
    \item Ice Cream Scoop - the reach of the thumb was too limited to press on the bar, so it was bent to allow for actuation.
\end{itemize}
The spray bottle and the thermometer were not modified in any way.





\bibliographystyle{IEEEtran} 
\bibliography{one_body_two_minds_arxive}

\end{document}